\documentclass[letterpaper, 10 pt, conference]{ieeeconf}  
\IEEEoverridecommandlockouts                              
\usepackage{graphics}
\usepackage{epsfig} 
\usepackage{textcomp}
\usepackage{times} 
\usepackage{amsmath} 
\usepackage{mathtools}
\usepackage{amssymb}  
\usepackage{url}
\usepackage{commath}
\usepackage{graphicx}
\usepackage{caption}
\usepackage{multirow}
\usepackage{subcaption}
\usepackage{array} 
\usepackage{url}
\usepackage{hyperref}
\hypersetup{colorlinks,allcolors=black}

\usepackage{siunitx}

\title{\LARGE \bf
A ROS-based Framework for Monitoring Human and Robot Conditions in a Human-Multi-robot Team 
}

\author{Wonse Jo, Shyam Sundar Kannan, Go-Eum Cha, Ahreum Lee, and Byung-Cheol Min
\thanks{Wonse Jo, Shyam Sundar Kannan, Go-Eum Cha, Ahreum Lee, and Byung-Cheol Min are with SMART Lab, Department of Computer and Information Technology, Purdue University, West Lafayette, IN 47907, USA \tt\small{jow@purdue.edu, kannan9@purdue.edu, cha20@purdue.edu, lahreum@purdue.edu, minb@purdue.edu}}%
}

\begin{document}
\maketitle
\thispagestyle{empty}
\pagestyle{empty}

\begin{abstract} 

This paper presents a framework for monitoring human and robot conditions in human multi-robot interactions.  The proposed framework consists of four modules: 1) human and robot conditions monitoring interface, 2) synchronization time filter, 3) data feature extraction interface, and 4) condition monitoring interface. The framework is based on Robot Operating System (ROS), and it supports physiological and behavioral sensors and devices and robot systems, as well as custom programs. Furthermore, it allows synchronizing the monitoring conditions and sharing them simultaneously. In order to validate the proposed framework, we present experiment results and analysis obtained from the user study where 30 human subjects participated and simulated robot experiments.

\end{abstract}

\section{INTRODUCTION}
\label{sec:introduction}
The recent advancements in robotics and communication have added up to the use of human-robot teams for various civilian and military applications. In these human-robot teams, the robots facilitate the performance of repetitive tasks over long hours with precision and accuracy. The role of a human is to provide the decision-making capabilities required to deal with unexpected situations. 
    But, it is uncommon that the intuition and the decision-making skills of the human operator are required all the time \cite{marks-2019}. 
    Since human intervention is not needed all the time, it is engaging to have human-multi-robot teams, where a human operator looks over multiple robots at a time. Lately, human-multi-robot teams are becoming more common. 
For example, in case of delivery robots, human in the loop assists the robots to complete arduous tasks like crossing the road and navigating in a crowded environment.

\begin{figure}[t]
    \centering
        \includegraphics[width=1\linewidth]{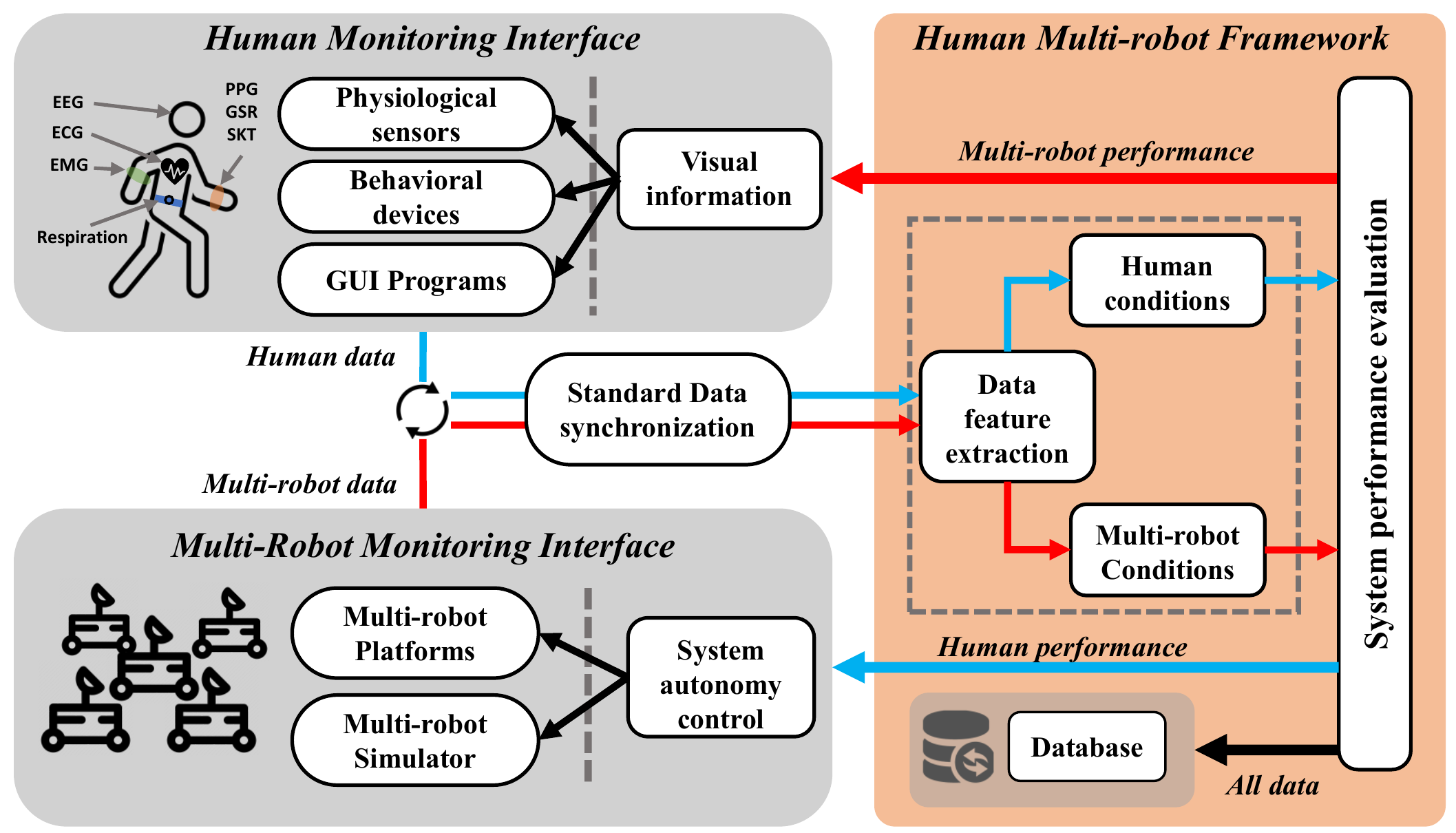}  
    \caption{An overview of the proposed ROS-based framework where the various human physiological and behavioural data and the robot conditions are monitored and shared simultaneously for human-multi-robot collaborations.} 
    \label{img:intro_image}
\end{figure} 

However, the existence of human is not always helpful to improve the overall performance of the system because it could be another variable to increase the complexity of the system \cite{humann2019human}.
The emotional and cognitive states of the human operator can cause human errors.
There are several variables that could affect their status 
such as the number of robots in the team, levels of autonomy of the robots, working duration, and operators trust in automation \cite{parasuraman1997humans}. Therefore, it is important to have adaptive systems that can reflect the state of the operator to dynamically adjust the level of autonomy and to re-allocate tasks to the robots. In this regard, physiological and behavioral data can be used as indicators of emotional and cognitive status of the human operator \cite{tiberio2013psychophysiological, lewis2010teams}. 

A framework to monitor the physiological data of human during human robot collaborative task was proposed in \cite{savur2019framework}. However, monitoring human conditions alone may not be sufficient to achieve a completely adaptive system \cite{bradshaw2003adjustable}. The adaptive system should reflect both the status of the robots and the human operator to enable dynamic task allocation and problem solving. 
Hence, it is necessary to build systems where the human 
and the robots conditions are monitored in parallel. 

\begin{figure*}[t]
    \centering
        \includegraphics[width=0.9\linewidth]{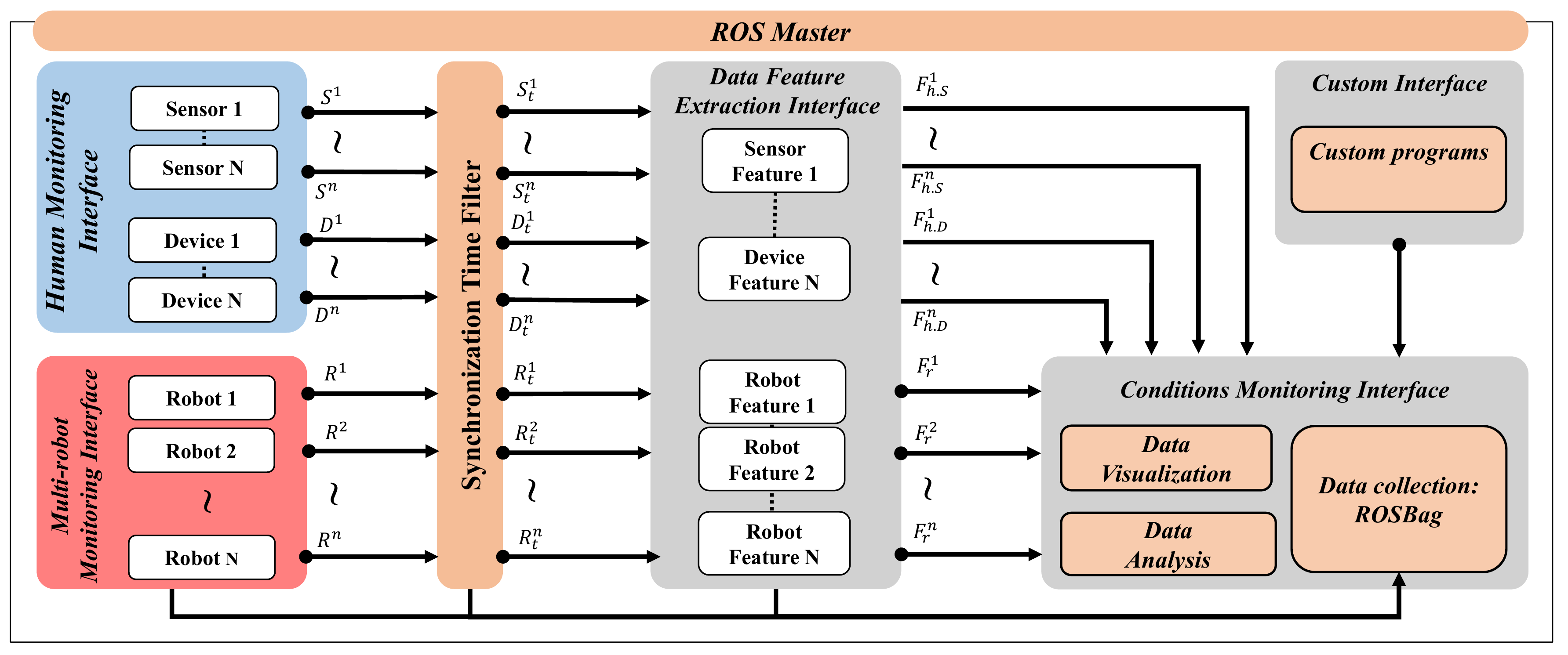}
    \caption{Generalized framework for monitoring the human and multi-robot conditions and sharing them simultaneously in a human-multi-robot team. The ROS-based framework integrates the sensory and device data of the human operator along with the robot operational data (from both real and simulated robots). The data from the various sources are synchronized using a time filter and then processed for feature extraction.} 
    \label{img:proposed_framework}
\end{figure*} 

In this paper, we present a framework for monitoring the human and the robot conditions that will play a vital role in adaptive human multi-robot systems that can dynamically respond to changes in situation. The framework is built upon Robot Operating System (ROS) \cite{quigley2009ros} to support various physiological and behavioral sensors and devices and robot systems. The framework constitutes three interfaces, one for monitoring the human and robot conditions, another for data feature extraction, and the other for data visualization. The human conditions are observed using physiological and behavioural monitoring. The robot conditions are monitored using various operation parameters of the robots such as battery level, internal temperature and functioning of the sensors. In addition to these, the framework sets up norms on the essential parameters for monitoring robot conditions. The frameworks also address the challenges and issues with synchronization of the data from various sources. In order to validate the proposed framework, user experiments and simulated robot experiments were conducted. An overview of the proposed framework has been illustrated in Fig. \ref{img:intro_image}.

The rest of the paper is organized as follows: Section \ref{sec:framework} presents the proposed framework and its components for monitoring human and robot states in the human-multi-robot team. Section \ref{sec:validation} elaborates on the validation of the proposed framework with human subjects and robots, followed by discussions in Section \ref{sec:discussion}. Conclusions are drawn with future work in Section \ref{sec:conclusion}.

\section{PROPOSED FRAMEWORK}
\label{sec:framework}
In this section, we present a ROS-based framework for monitoring the conditions of the human and the robots and sharing them simultaneously in a human multi-robot team. The overall architecture of the proposed framework has been depicted in Fig. \ref{img:proposed_framework}. The figure shows the various components of the proposed framework along with the data flow between its various components. The framework uses ROS as its underlying for all the interfacing and communications (rostopics used for all the communications). The framework supports various communication protocols like Bluetooth and WiFi, making it easy to seamlessly integrate the various physiological sensors and monitoring devices, as well as the multi-robot platforms and robotics simulator. The framework consists of the following modules: i) \textit{Monitoring interface}, ii) \textit{Synchronization time filter}, iii) \textit{Data feature extraction interface}, and iv) \textit{Condition monitoring interface}.

\subsection{Monitoring Interface}
The monitoring interface consists of a human monitoring interface and a multi-robot monitoring interface that are responsible for monitoring the human state and multi-robot states, respectively.
 
\subsubsection{Human Monitoring Interface}
The human state tends to vary with time, based on various factors such as workload, hours of operation, and so on. Hence, the human operator must be continuously monitored during human-robot tasks. Human vitals such as electrocardiogram (ECG), electroencephalogram (EEG), photoplethysmography (PPG), Inter Beat Interval (IBI), electromyography (EMG), and Galvanic Skin Response (GSR) tend to change based on the human's cognitive and emotional states \cite{ali2018globally}, \cite{kulic2007physiological}. These changes can be inferred through physiological monitoring using various physiological sensors. The human also tends to express his condition through external cues such as facial expressions, voice, and body gestures. These external behavioral patterns of the human operator must be also monitored alongside physiological monitoring. Hence in our framework, the Human Monitoring System consists of two sub-interfaces for physiological and behavioral monitoring.

The physiological sensor monitoring interface comprises various wearable physiological sensors $S^{n}$, $n\in \{1,..,N\}$ where $N$ is the number of sensors. These wearable devices monitor various human vitals such as ECG, EEG, GSR, EMG, and so on. Based on the trend and the variations in the physiological data, the human emotional and cognitive states can be estimated \cite{ali2018globally}, \cite{rohrmann1999changing}. 

The behavioral device monitoring interface uses various external devices $D^{n}$, $n\in\{1,..,N\}$ where $N$ is the number of devices, such as a camera, motion capture, and microphone to record the activities of the human operator. 

The analysis of both physiological and behavioral data using various machine learning techniques and computer vision can enable the estimation of the operator's state through these external cues \cite{nonis20193d}. 

\subsubsection{Multi-robot Monitoring Interface}
The multi-robot monitoring interface enables the connection of all the robots $R^{n}$, $n\in\{1,..,N\}$ where $N$ is the number of robots, to the system. The multi-robot team connected to the framework can either be a homogeneous or a heterogeneous team and can constitute both real and simulated robots. The framework being built upon ROS, inherits all the functionalities of it. The name-spacing feature which is fundamental in ROS is inherently used here to differentiate the data from the different robots.

\subsection{Synchronization Time Filter}
The framework integrating various $S^{n}$, $D^{n}$, and $R^{n}$ in real and/or simulator for monitoring humans and multi-robots are likely to be affected by synchronization issues. The various sensors used have different frequencies for publishing the data. Also, the sensors use different communication protocols and there can be differences in the latencies. In order to handle this, synchronization time filter is integrated into the framework. All the messages published ($S^{n}$, $D^{n}$, and $R^{n}$) include a timestamp, and then become $S^{n}_{t}$, $D^{n}_{t}$ and $R^{n}_{t}$. 
The timestamp is later used to synchronize the data. In addition to synchronization, the framework offers data storage using ROS bags and other visualization tools.

\subsection{Data Feature Extraction Interface}
The data feature extraction interface supports to add various feature extraction algorithm to extract features from the input data, such as $S^{n}_{t}$, $D^{n}_{t}$, and $R^{n}_{t}$. The output data ($F^{n}_{h,S}$, $F^{n}_{h,D}$ , and $F^{n}_{r}$) of this interface gives a better representation of the original dataset by reducing the redundant data. Hence, in physiology studies feature extraction is widely utilized to estimate human's emotion \cite{saganowski2019emotion,matsubara2016emotional}, stress \cite{hernando2015individual}, attention \cite{liu2013recognizing}, performance \cite{chanel2020towards} and so on. 

The feature extraction interface process the data from the multi-robot monitoring interface also yielding high level abstracted information such as:

\begin{itemize}
    \item \textit{Battery Utilization} is the measure of the current battery discharge rate of the robot. This measure helps to monitor whether the robot currently overloaded in terms of processing and resource utilization.
    \item \textit{Deployment Time} is the measure of the duration for which the robot has been actively deployed. 
    \item \textit{Sensor Status} given the life cycle of all the sensors on a robots. This helps in keeping a check whether all the sensors are functioning
\end{itemize}

\subsection{Condition Monitoring Interface}
The condition monitoring interface handles the collected data $\{S^{n}_{t}$, $S^{n}_{t}$, $F^{n}_{h.S}$, $F^{n}_{h.d}$, $F^{n}_{r}$\}, where $n \in \{1, ... , N\}$. This interface consists of three primary tools: data visualization, data analysis, and data collection. 

The data visualization tool enables simultaneous plotting of the received sensor and devices data using time series charts. In addition to the visualizing the sensory data, it also allows for the viewing of the  multiple video stream collected using the framework. 
 
The data analysis tools provide statistical analysis functionalities to perform data analysis and to find patterns by using statistical features, such as average, standard deviation, median, and so on. The framework allows for the extension of adding other analysis tools through the addition of other ROS node and packages.

The data collection tool saves all data synchronized with the ROS system as a ROSbag file. The ROSbag file enables the easily playback the collected data.

\section{Experiment}
\label{sec:validation}
We carried out experiments to collect data and test the functioning of the proposed framework. A user study was conducted to experiment the human condition monitoring interface of the framework and to build a ROSbag-based affective dataset, which includes physiological sensor data and behavioral device data. A simulated multi-robot experiment was also performed to test the multi-robot condition monitoring interface of the framework. Various robot operational data were collected. 

\subsection{Human Condition Monitoring Interface}


\begin{figure}
    \centering
     \begin{subfigure}[b]{1\linewidth}
        \centering 
        \includegraphics[width=0.95\linewidth]{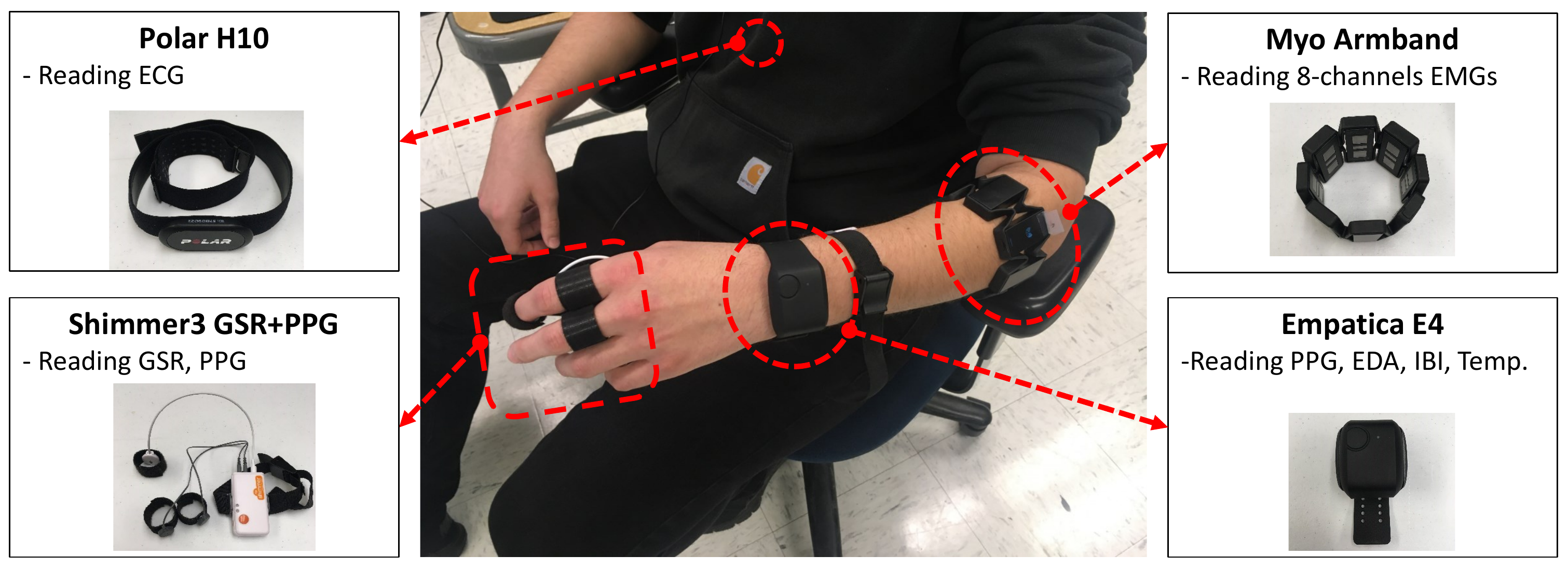} 
        \caption{Wearable Physiological monitoring sensors}
        \label{img:supporting_sensors}
    \end{subfigure}
    
    \begin{subfigure}[b]{1\linewidth}
        \centering 
        \includegraphics[width=0.95\linewidth]{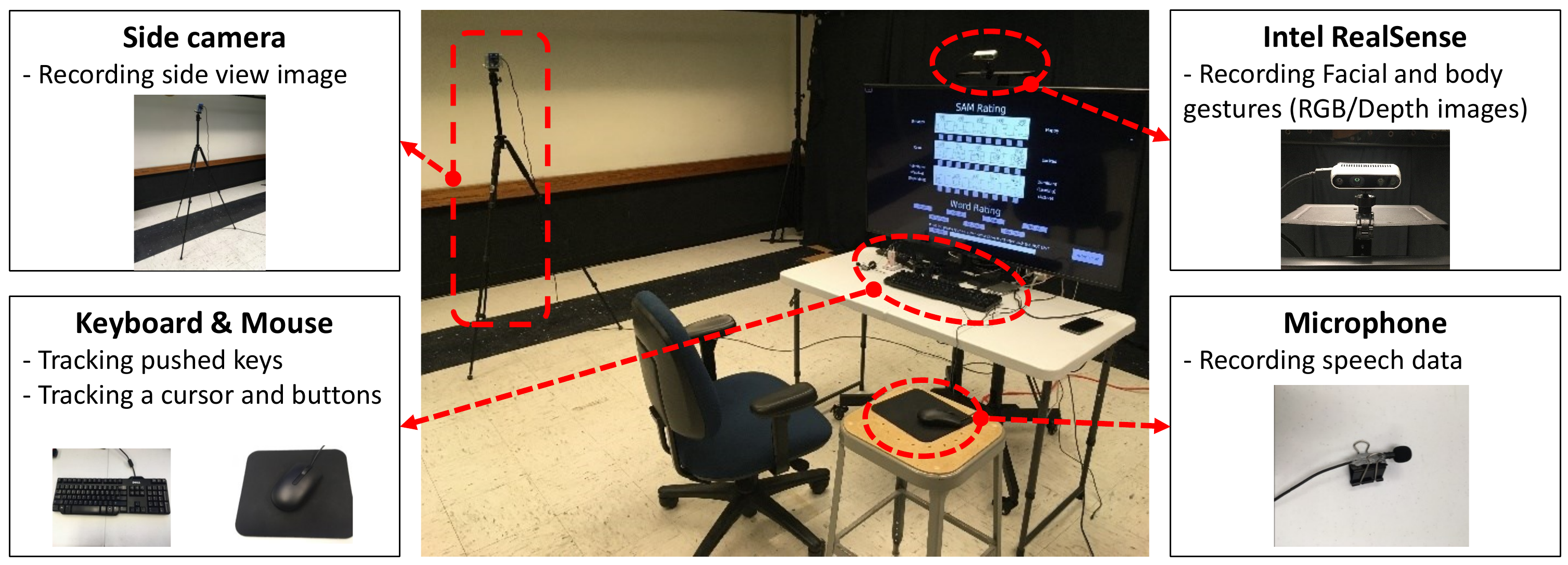} 
        \caption{Behavior monitoring devices}
        \label{img:supporting_devices}
    \end{subfigure}
    \caption{Hardware and sensors used in the user study; (a) the physiological monitoring sensors worn by the user, and (b) the behavioral monitoring devices placed in the experimental setup.}
    \label{img:supporting_sensors_devices}
\end{figure}

The human condition monitoring interface of the proposed framework was experimented via a user study where 30 human subjects participated (19 males and 11 females; average age: 25.1). This study was approved by the University's Institutional Review Board (Purdue IRB Protocol: \#1812021453).

The participants were given two tasks, affective and cognitive tasks. For the affective task, participants were asked to look at images and then listen to short audio clips. The visual and auditory stimuli were given to alter the human emotional state. A controlled experiment was performed so as to stimulate various human emotions so that the ability of the framework to record various human emotions could be tested. The images and the audio clips were taken from the International Affective Picture System (IAPS) \cite{lang2005international} and the International Affective Digitized Sound System (IADS) \cite{bradley2007international} which are widely used and validated in the physiology filed for eliciting specific emotions \cite{sanchez2018artificial, hsu2018affective}. For the cognitive task, the participants were given the dual n-back game \cite{hampson2006brain} to elicit different levels of the workload. Dual $n$-back game is a brain training exercise where the user needs to match the current position and audio information (sound of an English alphabet) to ones shown and played $n$-steps before. In the experiment, the participants were given dual 1-back, dual 2-back and then, dual 3-back. The number of back steps was increased in every iteration so as to increase the workload imposed. Each round lasted for $60$ seconds and all the human data were recorded using the framework.

\begin{figure}
    \centering
        \includegraphics[width=1\linewidth]{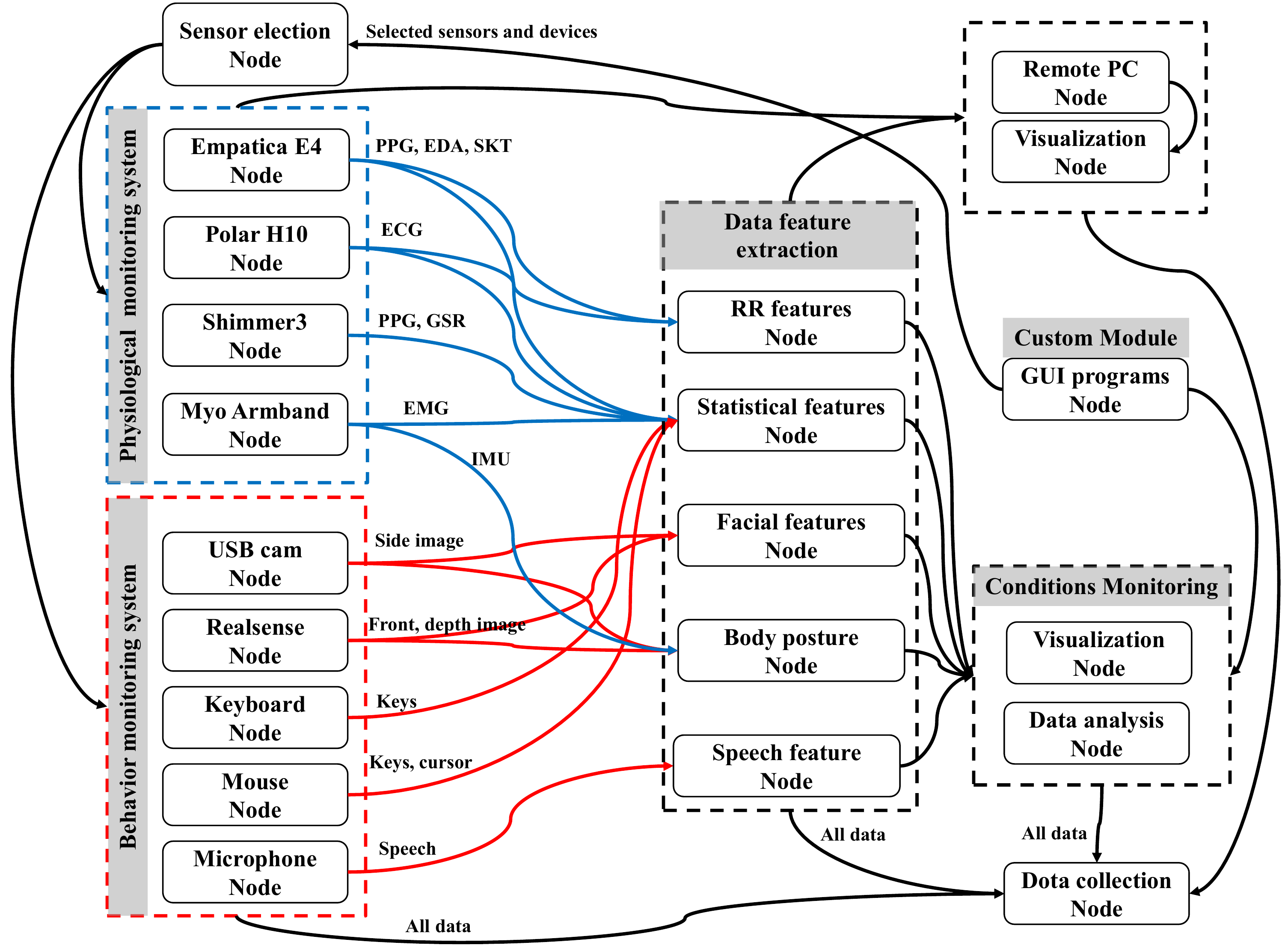}
    \caption{ROS node graph utilized on the user experiment.} 
    \label{img:summary_node}
\end{figure}

During the given two tasks, participants were asked to wear the various wearable sensors, while other external sensors were monitoring the participants, as shown in Fig. \ref{img:supporting_sensors_devices}. In this study, we tried to use as many sensors as possible to obtain human response data from the tasks.
Empatica E4 \cite{empatica}, Myo \cite{myoarmband}, Polar H10 \cite{polarusa}, and Shimmer3 GSR+ \cite{shimmer} were used to monitor physiological responses of the participants. Behavioral responses were monitored with an Intel Realsense \cite{intelrealsense}, a side camera, and input devices such as a mouse, a keyboard, and a microphone. 
All the responses of each test were recorded as a ROSbag file of the proposed framework, which enables further analysis of user study. The Fig. \ref{img:summary_node} shows the details of ROS node used in this user study.

\subsection{Multi-Robot Condition Monitoring Interface}
\begin{figure}[t]
    \centering
        \includegraphics[width=0.8\linewidth]{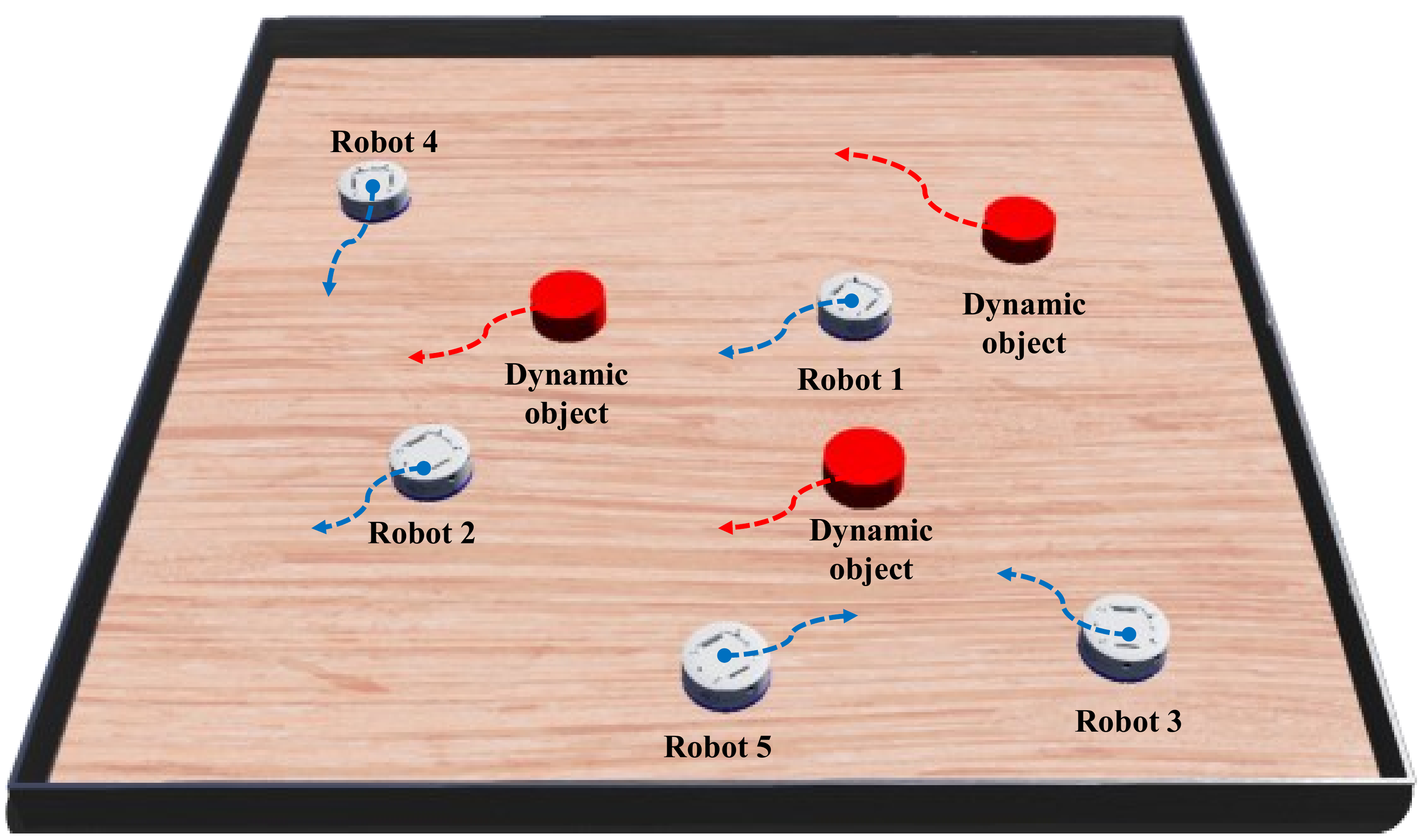} 
    \caption{Webots simulation setup used in the experiment to test the multi-robot monitoring interface.}
    \label{img:simulation_setting}
\end{figure}

In order to validate the robot condition monitoring interface of the framework, simple multi-robot experiments were conducted using Webots simulator \cite{michel2004cyberbotics}. In the experiment, five mobile robots and three dynamic obstacles were employed, and the battery condition of the robots and the wireless signal strength between the robots and the router were monitored and recorded. A battery is directly related to the robot's operating time, and wireless signal strength is directly related to communication quality, which are often used to monitor the robot's condition. To this end, a WiFi router was installed at the lower left corner of the workspace (boxes) of the simulation environment, and robots were wirelessly connected to the router. All the five robots were initially placed close to the top right corner of the workspace. The experiment setup used is depicted in Fig. \ref{img:simulation_setting}. During the experiment, the robots were programmed to perform random walks while avoiding all the obstacles. In order to simulate fluctuations in the battery utilization, we changed the CPU consumption parameter in the simulator during the experiment, which is related to the battery consumption.

\begin{figure}
    \centering
        \includegraphics[width=0.95\linewidth]{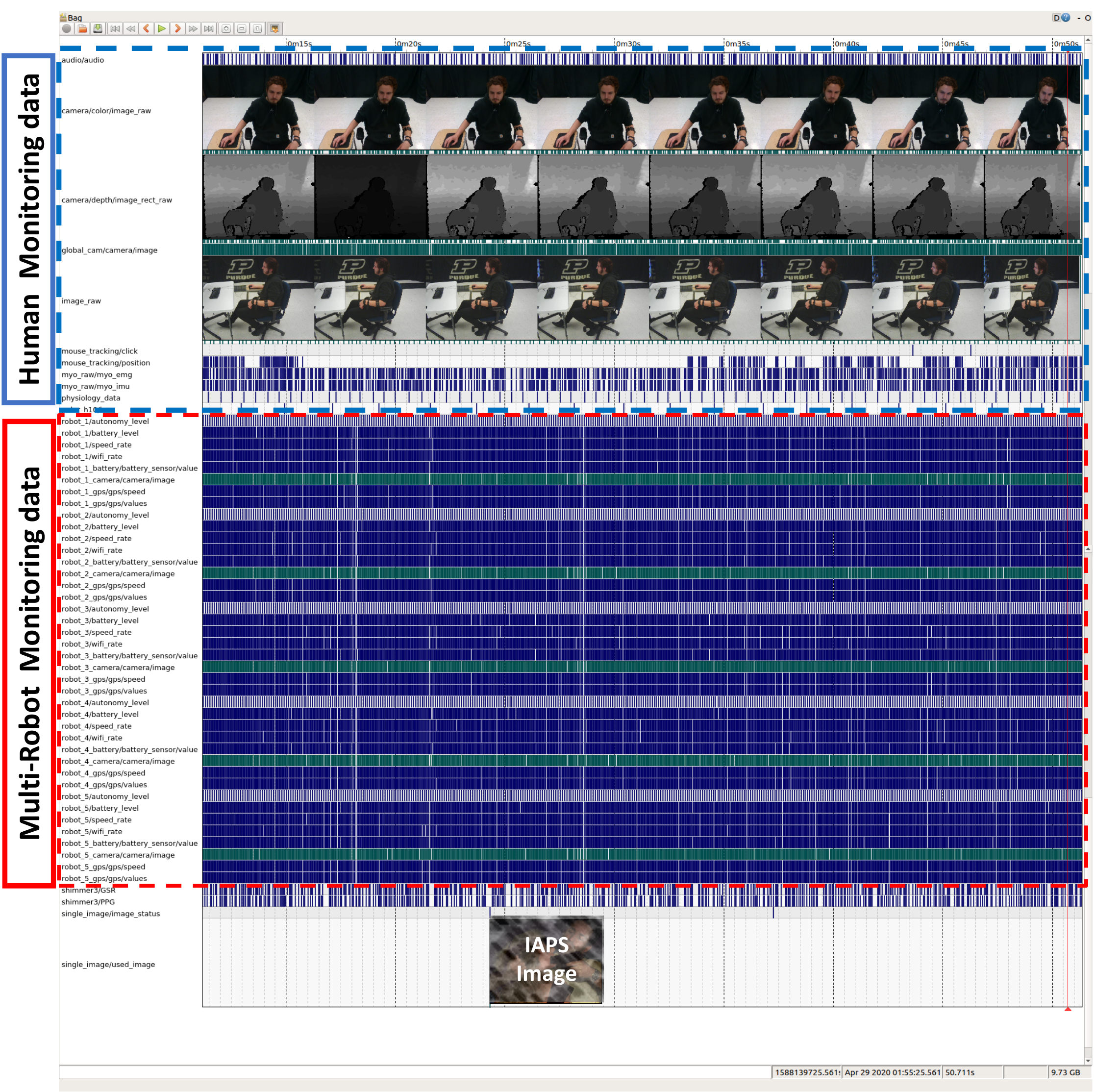} 
    \caption{Sample of the synchronized dataset via ROS system.}
    \label{img:data_collection}
\end{figure}

\begin{figure*}[t]
    \centering
     \begin{subfigure}[b]{0.49\linewidth}
        \centering 
        \includegraphics[width=1\linewidth]{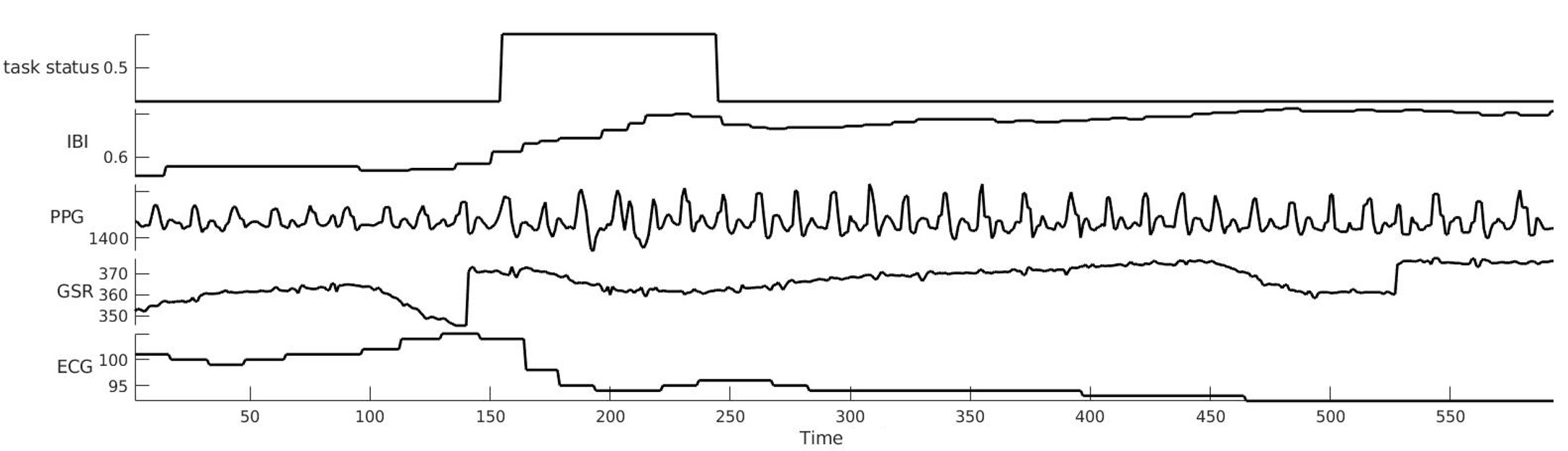} 
        \caption{Plot of the data collected with image as stimulus.}
        \label{img:data_emotion}
    \end{subfigure}
    \begin{subfigure}[b]{0.49\linewidth}
        \centering 
        \includegraphics[width=1\linewidth]{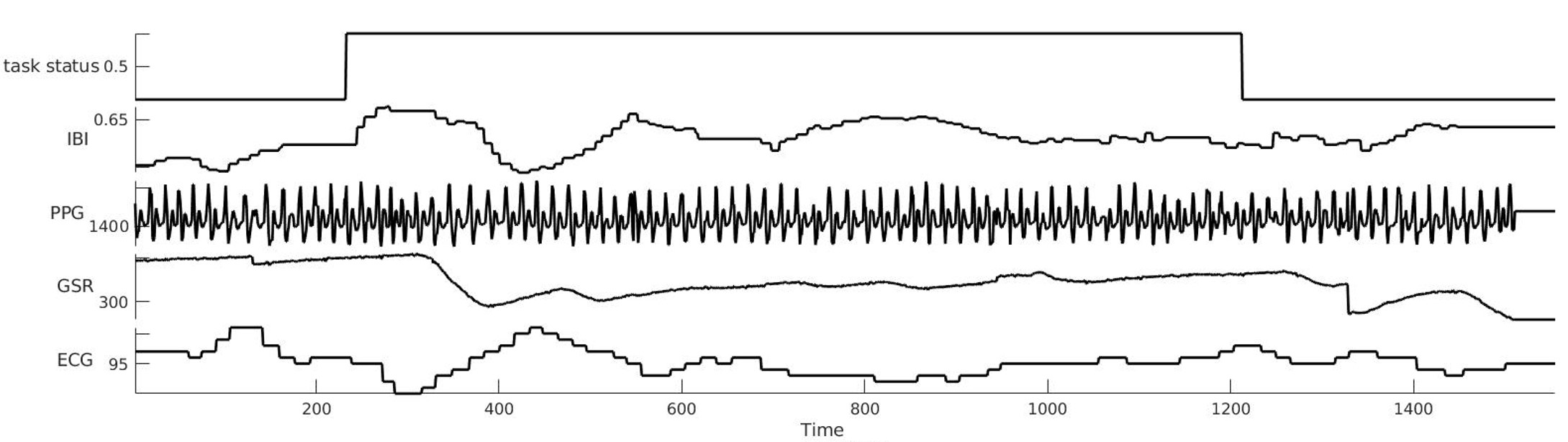} 
        \caption{Plot of the data collected during dual $1$-back task.}
        \label{img:dataset_dual_1_back}
    \end{subfigure}
    
    \begin{subfigure}[b]{0.49\linewidth}
        \centering 
        \includegraphics[width=1\linewidth]{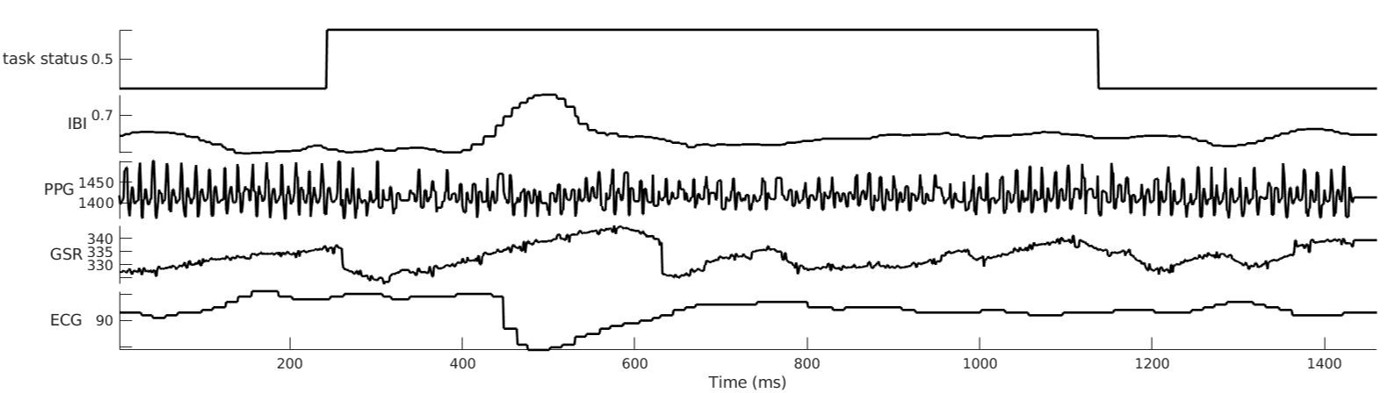} 
        \caption{Plot of the data collected during dual $2$-back task.}
        \label{img:dataset_dual_2_back}
    \end{subfigure}
    \begin{subfigure}[b]{0.49\linewidth}
        \centering 
        \includegraphics[width=1\linewidth]{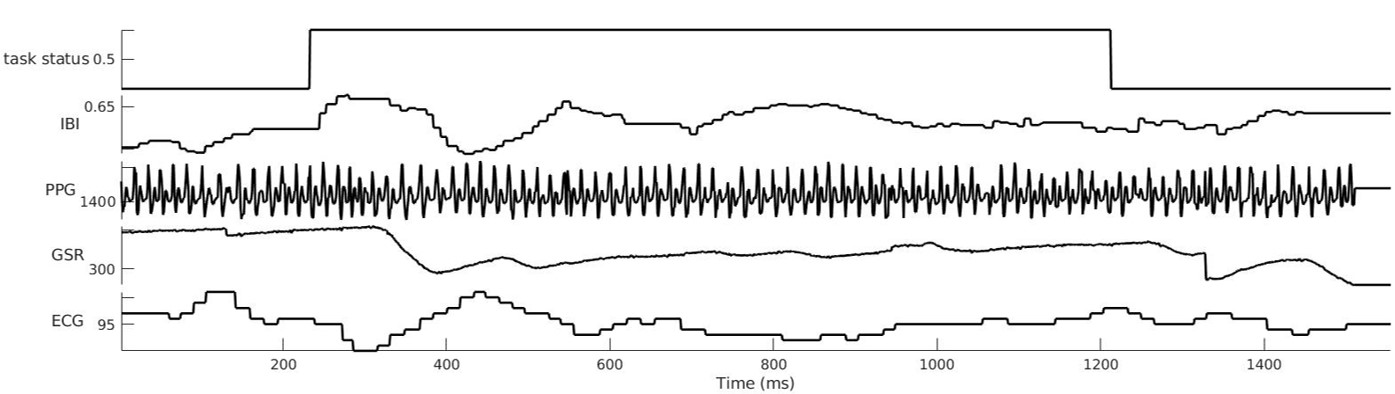} 
        \caption{Plot of the data collected during dual $3$-back task.}
        \label{img:dataset_dual_3_back}
    \end{subfigure}
    
    \caption{Physiological data collected from a participant during the emotion (image stimulus) and workload tasks.}
    \label{img:physcholgy_data}
\end{figure*}

\begin{figure*}[t]
    \centering
     \begin{subfigure}[b]{0.24\linewidth}
        \centering 
        \includegraphics[width=1\linewidth]{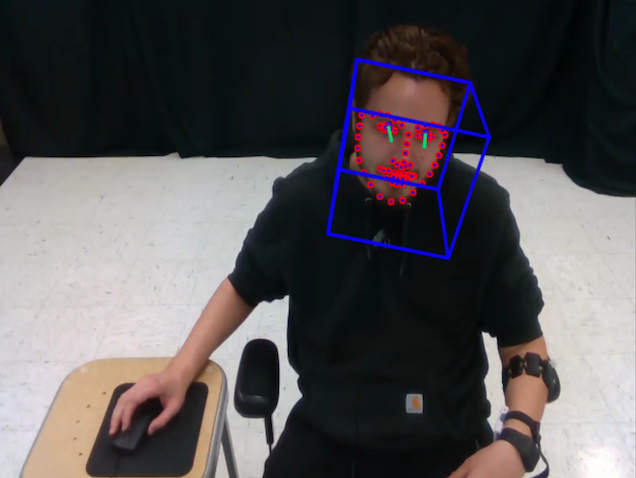} 
        \caption{Extracted facial features from user study data}
        \label{img:validation_openface}
    \end{subfigure}
    \begin{subfigure}[b]{0.24\linewidth}
        \centering 
        \includegraphics[width=1\linewidth]{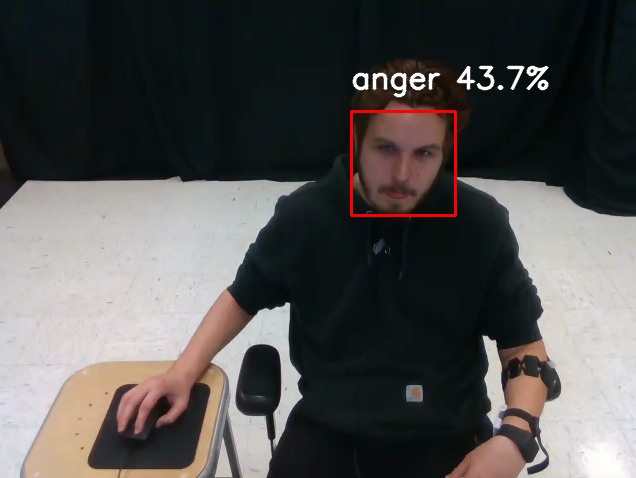} 
        \caption{Emotion detection from user study data}
        \label{img:validation_emopy}
    \end{subfigure}
    \begin{subfigure}[b]{0.24\linewidth}
        \centering 
        \includegraphics[width=1\linewidth]{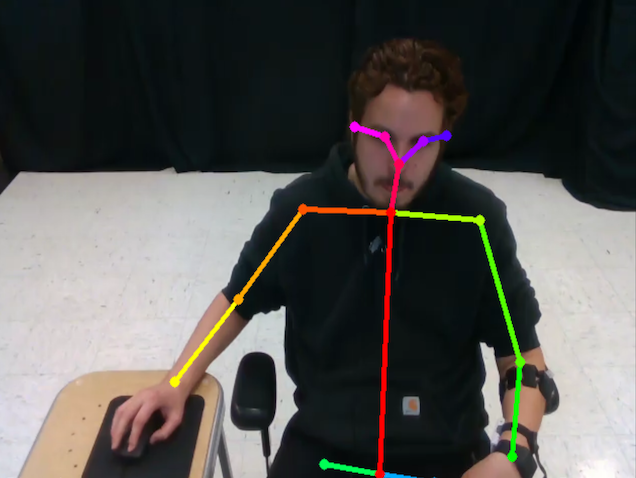} 
        \caption{Body posture extracted from the Intel Realsense camera.}
        \label{img:validation_openpose}
    \end{subfigure}
    \begin{subfigure}[b]{0.24\linewidth}
        \centering 
        \includegraphics[width=1\linewidth]{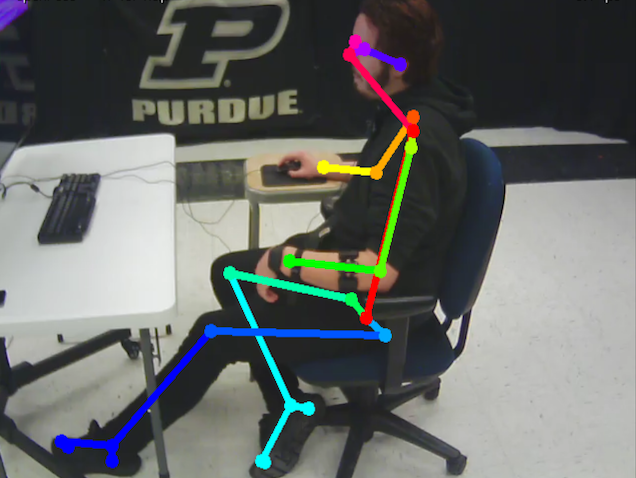} 
        \caption{Body posture extracted from the side camera.}
        \label{img:validation_openpose_side}
    \end{subfigure}
    \caption{Analysis examples of behavioral features; (a) Facial features by OpenFace 2.0 \cite{baltrusaitis2018openface}, (b) Emotion detection by Emopy \cite{thoughtworksarts2020}, (c) Body posture from the Intel Realsense camera by OpenPose \cite{cao2018openpose} and (d) Body posture from the side camera by Openpose \cite{cao2018openpose}}
    \label{img:validation}
\end{figure*}

\section{Validation and Results}
\label{sec:discussion}
In this section, the data collected during the experiments from both the humans and the robot using the proposed framework are validated and further analyzed as shown Fig. \ref{img:data_collection}. The data collected were validated in terms of their synchronization and reliability towards monitoring the changes in the human and the robots conditions. 

\subsection{Validation of Human Condition Monitoring Interface}

\subsubsection{Analysis of Physiological Data }
During the experiments, various physiological vitals such as ECG, IBI, EMG, PPG, and GSR were collected from the various wearable sensors. These data were collected from devices which publish data at various frequencies. Also, the devices use various communication protocols such a Bluetooth Low Energy and WiFi, which can have varying latencies. To validate the synchronization of the data collected, the fluctuations in the data were examined. 

Data from individual sensors were compared with the timeline of the experiment. From the comparison of the data with the experiments timeline, a sudden fluctuation in the data were observed almost at the same when the experimental stimulus was introduced. Fig. \ref{img:data_emotion} shows the IBI, PPG, GSR and the ECG data collected from a participant when an image stimulus was shown. It can be observed that there is a sudden deviation in the trends of the IBI, GSR and the ECG data when the image was introduced to the participant (shown by the task status). It can be seen that these swings happens at the same time and hence validating that the synchronization time filter was able to the synchronize the data despite the latencies and different functioning frequencies of the sensors. 

Then, based on the sensory data collected during the workload task, a trend in the data was observed with the increase in the workload from dual $1$-back to dual $3$-back. The IBI, PPG, GSR and the ECG collected during the workload task from a participant are shown in Fig. \ref{img:dataset_dual_1_back}, \ref{img:dataset_dual_2_back} and \ref{img:dataset_dual_3_back} along with the mean, standard deviation and median of the difference between the baseline and recorded data in Table. \ref{tab:dataset_anaysis}. From the table, it can be seen that there is a trend average values. For instance, the mean IBI values increases with the increase in the workload from dual $1$-back to dual $3$-back. The average ECG value looks to decrease with the tasks. These trends substantiate that the framework can monitor the changes in the data and they are in coherence with the humans state. This adds to the reliability of the framework towards human condition monitoring.

\begin{table}[t]
\centering
\caption{Average, standard deviation (S.D.) and median of the difference between the baseline and recorded data for the workload tasks.}
\label{tab:dataset_anaysis}
\begin{tabular}{cccccc}
\hline
 &
  Features &
  \begin{tabular}[c]{@{}c@{}}IBI\end{tabular} &
  \begin{tabular}[c]{@{}c@{}}PPG\end{tabular} &
  \begin{tabular}[c]{@{}c@{}}GSR\end{tabular} &
  \begin{tabular}[c]{@{}c@{}}ECG\end{tabular} \\ \hline
\multirow{3}{*}{\begin{tabular}[c]{@{}c@{}}Dual\\ 1-back\end{tabular}} & Average & -0.0329 & 3.1639  & -0.1807  & 2.3941  \\
                                                                       & S.D.    & 0.0039  & 3.2211  & 4.6100   & 1.0129  \\
                                                                       & Median  & -0.0422 & 2.5641  & 1.7990   & 4.0000  \\ \hline
\multirow{3}{*}{\begin{tabular}[c]{@{}c@{}}Dual\\ 2-back\end{tabular}} & Average & 0.0176  & -3.2948 & 1.6718   & -2.2991 \\
                                                                       & S.D.    & 0.0134  & -8.4760 & 2.3872   & 2.0459  \\
                                                                       & Median  & 0.0203  & -3.6630 & 0.6480   & -1.0000 \\ \hline
\multirow{3}{*}{\begin{tabular}[c]{@{}c@{}}Dual\\ 3-back\end{tabular}} & Average & 0.0229  & -0.5907 & -26.5982 & -2.5881 \\
                                                                       & S.D.    & 0.0059  & -0.0818 & 11.3487  & 0.4228  \\
                                                                       & Median  & 0.0219  & 0.7326  & -28.7434 & -2.0000 \\ \hline
\end{tabular}
\end{table}

\subsubsection{Analysis of Behavioural Data }
Further, the visual data collected from the Intel Realsense depth camera and the side camera during the user experiment were analyzed to extract information such as human postures, facial expressions and emotions. From the 3D facial image obtained from the Intel Realsense camera, we were able to extract the facial features using OpenFace 2.0 \cite{baltrusaitis2018openface} which are essential for further analysis of the facial expressions. The facial features are extracted as shown in Fig. \ref{img:validation_openface}. It can be seen that all the facial features such as eyes, mouth, nose and facial boundary are extracted precisely. The emotion of the user was obtained using EmoPy \cite{thoughtworksarts2020} and the result of the estimated emotion along with the confidence has been shown in Fig. \ref{img:validation_emopy}. Further analysis on the postures were done using Openpose \cite{cao2018openpose} on visual data obtained from both the Intel Realsense and the side camera. The body postures were extracted accurately from both the cameras. The results of the body pose estimated are shown in Fig. \ref{img:validation_openpose} and Fig. \ref{img:validation_openpose_side}.

From these extracted features and the pose information, it can be seen that the proposed framework enables the extraction of desired high level information. The facial emotion extracted also matched with the nature of the stimulus used in the user experiment. The facial features and the body gestures extracted opens gate for further research on processing these information for estimating the humans state. Hence, validating the ability of the framework to monitor the behavioural aspects of the human.   

\subsection{Analysis of Multi-Robot Condition Monitoring Interface}
\begin{figure}[t]
    \centering
    \includegraphics[width=1\linewidth]{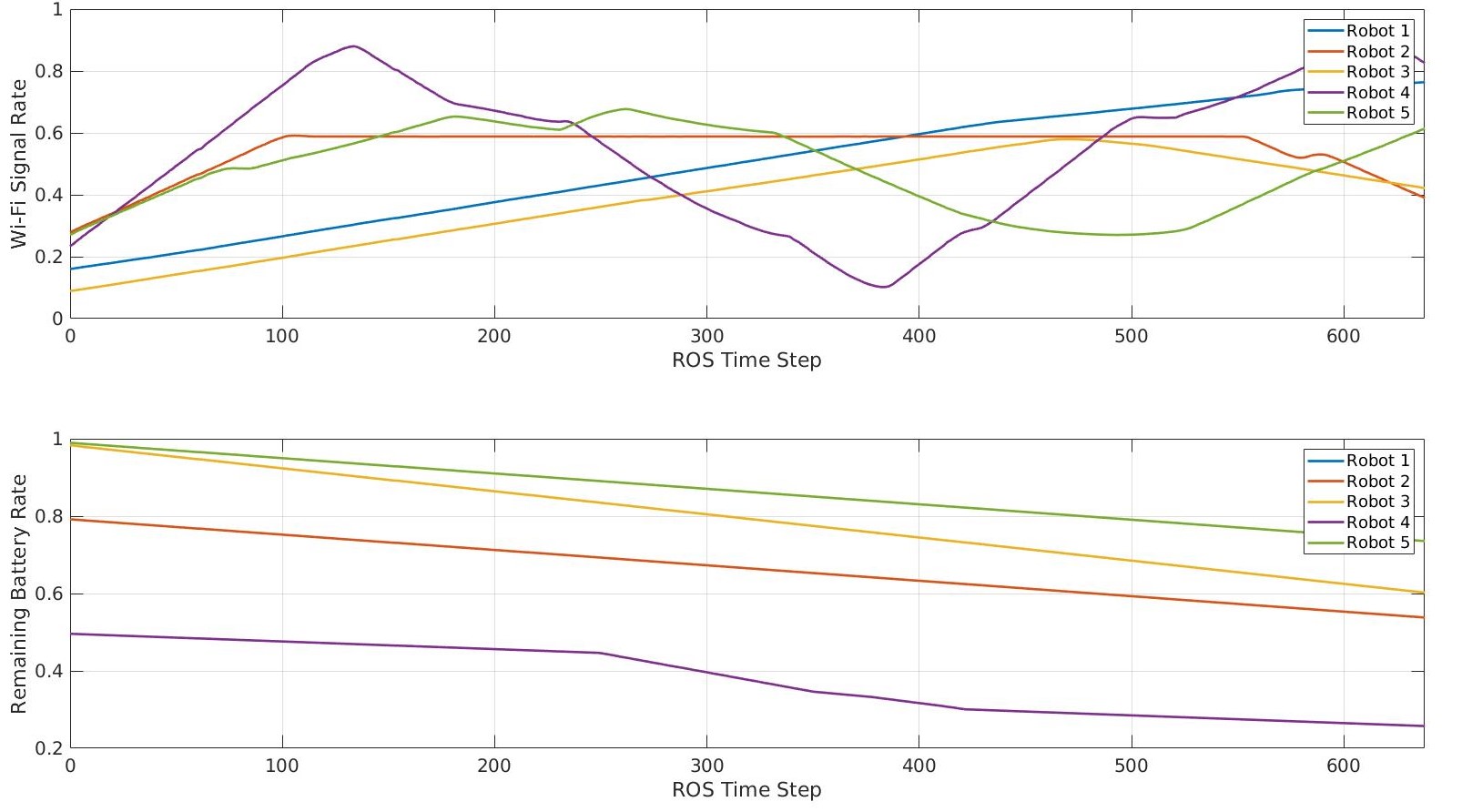} 
    \caption{WiFi signal strength (top) and remaining battery (bottom) of the five robots recorded using the multi-robot monitoring interface.}
    \label{img:simulator_result_data}
\end{figure}

Similar to the human condition monitoring interface, the data from the multi-robot monitoring interface were also analyzed. The WiFi signal strength and remaining battery data collected during the experiment are illustrated in Fig. \ref{img:simulator_result_data}. From the figure it can be seen that the WiFi signal strengths were weak at the beginning of the experiment. This is coherent with the fact that the robots were initially placed at the other extreme end from the location of the router. During the experiment, the robots moved around in the environment while avoiding obstacles, and this resulted that their signal strengths changed with time. This can be seen in Fig. \ref{img:simulator_result_data} (top). Also, a fluctuation in the battery utilization rate of the robot 4 can be observed between ROS time step 250 to 420 as shown in Fig. \ref{img:simulator_result_data} (bottom). This is synchronous with the change in the CPU consumption parameter of the robot 4 amidst the experiment. This validates the ability of the framework to monitor the fluctuations in the multi-robot conditions.

\section{CONCLUSION and FUTURE WORKS}
\label{sec:conclusion}
In this work, a ROS-based framework has been proposed for monitoring the states of the human and the robots in a human-multi-robot team. The objective of this framework is to observe states of the human and conditions of robots and to analyze the collected data for further investigation. This framework allowed us to observe the physiological and behavioral responses of operators with diverse wearable sensors and behavioral monitoring devices, which is important to build an enhanced adaptive system for a human-multi-robot team. 

Not only do humans play an essential role in human-robot tasks, but also robots do. The framework monitored the functional capabilities of robots. Since the factors of robots, such as battery utilization, deployment time, and sensor status, can affect the employment of robots directly, measuring robot capacities was implemented in the framework as well. Taking advantage of ROS that can be easily integrated with the robots, the framework was able to share human and robot states simultaneously. A user study and a robot simulation validated the proposed framework from the result of analysis on collected data. 

In the future, we intend to perform a full scale experiment where a human operator controls a team of multiple robots. Another potential direction of the future works is to develop adaptive control systems where the level of robot's autonomy and the task allocated to the human operator can be dynamically adjusted based on their conditions. This ensures that the both the human and the robot are not forced to function beyond their capabilities.



\section*{ACKNOWLEDGMENT}
This work was supported by NSF CAREER Award IIS-1846221.

\bibliography{root}
\bibliographystyle{IEEEtran}
\end{document}